\documentclass[letterpaper]{article} 

\ifdefined\aaaianonymous
    \usepackage[submission]{aaai2026}  
\else
    \usepackage{aaai2026}              
\fi

\usepackage{times}  
\usepackage{helvet}  
\usepackage{courier}  
\usepackage[hyphens]{url}  
\usepackage{graphicx} 
\usepackage{natbib}  
\usepackage{caption} 
\usepackage{algorithm}
\usepackage{algorithmic}

\usepackage{newfloat}
\usepackage{listings}
\DeclareCaptionStyle{ruled}{labelfont=normalfont,labelsep=colon,strut=off} 
\floatstyle{ruled}
\newfloat{listing}{tb}{lst}{}
\floatname{listing}{Listing}

\usepackage{amsmath}
\usepackage{amssymb}
\usepackage{tikz}
\usepackage{subcaption}
\usepackage{xcolor}
\usepackage{multirow}
\usepackage{booktabs}

\ifdefined\aaaianonymous
    \title{Learning to Optimize Job Shop Scheduling Under Structural Uncertainty}
\else
    \title{Learning to Optimize Job Shop Scheduling Under Structural Uncertainty}
\fi

\author{
    Rui Zhang\textsuperscript{\rm 13},
    Jianwei Niu\textsuperscript{\rm 123}\thanks{Corresponding Author.},
    Xuefeng Liu\textsuperscript{\rm 13}\footnotemark[1], 
    Shaojie Tang\textsuperscript{\rm 4},
    Jing Yuan\textsuperscript{\rm 5}
}
\affiliations{
    \textsuperscript{\rm 1}State Key Laboratory of Virtual Reality Technology and Systems, School of Computer Science \\and Engineering, Beihang University, Beijing, China\\
    \textsuperscript{\rm 2}Hangzhou Innovation Institute of Beihang University, \\Zhejiang Key Laboratory of Industrial Big Data and Robot Intelligent Systems, Hangzhou, China\\
    \textsuperscript{\rm 3}Zhongguancun Laboratory, Beijing, China\\
    \textsuperscript{\rm 4}Department of Management Science and Systems, University at Buffalo, Buffalo, New York, USA\\
    \textsuperscript{\rm 5}University of North Texas, Denton, Texas, USA\\
    \{gary\_zhang, niujianwei, liu\_xuefeng\}@buaa.edu.cn,
    shaojiet@buffalo.edu,
    jing.yuan@unt.edu
}

\usepackage{bibentry}

\begin{document}

\maketitle

\begin{abstract}
The Job-Shop Scheduling Problem (JSSP), under various forms of manufacturing uncertainty, has recently attracted considerable research attention. Most existing studies focus on parameter uncertainty, such as variable processing times, and typically adopt the actor-critic framework. In this paper, we explore a different but prevalent form of uncertainty in JSSP: structural uncertainty. Structural uncertainty arises when a job may follow one of several routing paths, and the selection is determined not by policy, but by situational factors (e.g., the quality of intermediate products) that cannot be known in advance. Existing methods struggle to address this challenge due to incorrect credit assignment: a high-quality action may be unfairly penalized if it is followed by a time-consuming path. To address this problem, we propose a novel method named UP-AAC. In contrast to conventional actor-critic methods, UP-AAC employs an asymmetric architecture. While its actor receives a standard stochastic state, the critic is crucially provided with a deterministic state reconstructed in hindsight. This design allows the critic to learn a more accurate value function, which in turn provides a lower-variance policy gradient to the actor, leading to more stable learning. In addition, we design an attention-based Uncertainty Perception Model (UPM) to enhance the actor's scheduling decisions. Extensive experiments demonstrate that our method outperforms existing approaches in reducing makespan on benchmark instances.
\end{abstract}


\section{Introduction}

The Job-Shop Scheduling Problem (JSSP) is a classic NP-hard optimization challenge, fundamental to efficient operations in manufacturing and logistics \citep{Gao2019ARO, Mao2019OpportunitiesAC}. The primary goal is to schedule a set of jobs on various machines to minimize the total completion time, or makespan. Due to its complexity, traditional approaches often rely on heuristic methods, such as Priority Dispatching Rules (PDRs) \citep{Haupt1989ASO}. Recently, a new paradigm has emerged by combining Deep Reinforcement Learning (DRL) with Graph Neural Networks (GNNs), often within an Actor-Critic (AC) framework \citep{Smit2024GraphNN, ho2024residual}. This methodology has demonstrated remarkable success on standard deterministic JSSP benchmarks, creating agents that outperform many traditional heuristics.

\begin{figure}[t]
    \centering
    \includegraphics[trim=0pt 4pt 0pt 12pt, clip, width=\columnwidth]{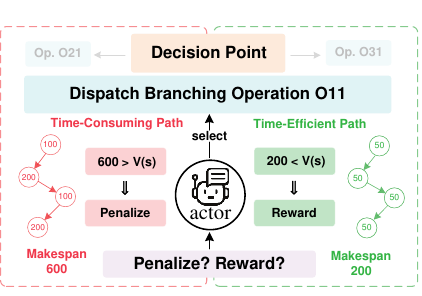}
    \caption{An example of the credit assignment problem. The same dispatched operation can lead to drastically different makespans due to random path realizations. This unfairly rewards or penalizes the actor's choice, obscuring the action's true quality and hindering the learning process.}
    \label{fig:toy_example}
\end{figure}

However, the deterministic assumption presents a significant limitation, as real-world manufacturing is inherently uncertain. In response, a key research direction has been the extension of DRL to handle such uncertainty. Most existing work has focused on parameter uncertainty (e.g., variable processing times) \citep{infantes2024learning} or dynamic events (e.g., machine breakdowns, new job arrivals) \citep{luo2020dynamic, Lei2024},  typically by adapting the standard AC framework. In this paper, we focus on a distinct yet equally critical challenge: structural uncertainty. This form of uncertainty emerges when a job’s processing route is not fixed but may branch into multiple paths, with the actual path determined dynamically by situational factors. Such scenarios are common in complex industrial settings. For instance, the fabrication path of a silicon wafer may change based on intermediate quality inspections, while the testing sequence of a biological sample might depend on initial screening results. These conditions transform the scheduling problem from a static sequencing task into a dynamic planning challenge under profound structural uncertainty.

The presence of structural uncertainty poses a specific challenge to the standard AC framework: the issue of incorrect credit assignment. As illustrated in Figure~\ref{fig:toy_example}, the evaluation of an action becomes strongly coupled with the stochastic path that unfolds after the action is taken, rather than reflecting the action’s intrinsic quality. Specifically, a standard Critic learns to estimate the average outcome over all possible future paths. As a result, it may unfairly penalize a high-quality action that is followed by a time-consuming path, or reward a poor action that happens to be followed by a time-efficient one. This high variance, driven by environmental stochasticity rather than the policy’s actual performance, undermines the reliability of the Critic’s baseline, resulting in unstable gradients and ineffective policy optimization.

To address this challenge, we propose UP-AAC, an integrated DRL framework with two key innovations. At its core is a novel Asymmetric Actor-Critic (AAC) architecture. Unlike in standard AC where the Critic observes the same uncertain states as the Actor \citep{smit2025neural, zhang2020learning}, our AAC redefines the Critic's learning process by training it on a deterministic hindsight state, reconstructed using the specific path actually realized during an episode. This approach allows the Critic to evaluate an action against its true, realized consequences, rather than on a noisy average over all possibilities.  By removing variance introduced by environmental uncertainty, this design directly resolves the credit assignment problem and provides a stable, reliable learning signal for effective policy optimization. Complementing this core architecture, our Uncertainty Perception Model (UPM) provides the Actor with a high-level summary of the problem's overall uncertainty, enabling it to make more robust and forward-looking decisions.

In summary, our key contributions are:

\begin{itemize}
\item We propose UP-AAC, a novel DRL framework to solve JSSP with structural uncertainty, featuring a core Asymmetric Actor-Critic architecture that enables stable learning in highly stochastic environments.
\item We introduce UPM, a knowledge-guided module that explicitly quantifies and incorporates uncertainty, enhancing policy robustness.
\item We conduct extensive experiments and show that our method achieves state-of-the-art performance, outperforming a wide range of baselines.
\end{itemize}

\section{Related Work}

Neural network approaches to the JSSP can be divided into two main streams: deterministic scheduling methods and uncertainty-aware methods.

\subsubsection{Deterministic Scheduling Methods}

Most research in deterministic scheduling employs DRL with GNN to learn constructive heuristics. A pioneering GNN-based policy demonstrated strong generalization \citep{zhang2020learning, park2021learning}, leading to extensive follow-up work. One research direction focused on architectural enhancements, incorporating attention mechanisms or Transformers to model complex dependencies \citep{yang2022using, Chen2023a, Lee2024, zhang2024novel}, as well as specialized structures like dual-attention and heterogeneous GNNs for the more complex Flexible JSP (FJSP) \citep{Wang2023, zhao2024dual, song2022flexible, tang2024solving}. Another direction explored innovative learning paradigms, such as learning improvement heuristics \citep{zhang2022deep}, self-supervised methods \citep{Corsini2024SelfLabelingTJ, Pirnay2024, Pirnay2024a}, regret-based training \citep{sun2024learning}, and non-DRL approaches like Lagrangian dual deep learning \citep{kotary2022fast}. A key limitation of these methods is their assumption of a static, deterministic environment.

\subsubsection{Uncertainty-aware Methods}

To address uncertainty, research has evolved in several directions. For dynamic scheduling with events like new job arrivals or machine failures, DRL has been used to develop adaptive, real-time policies \citep{luo2020dynamic, han2020research, Liu2022, Luo2022, Lei2024}. This work, however, primarily handles external disruptions. Another category deals with parameter uncertainty, mainly by extending DRL to manage stochastic processing times \citep{su2023evolution, zhang2023deep, wu2024deep, liu2024dynamic}. Notably, some research has focused on creating robust policies by explicitly learning uncertainty representations from multiple sampled scenarios \citep{smit2025neural}.

\begin{figure*}[h!]
    \centering
    \includegraphics[width=\textwidth]{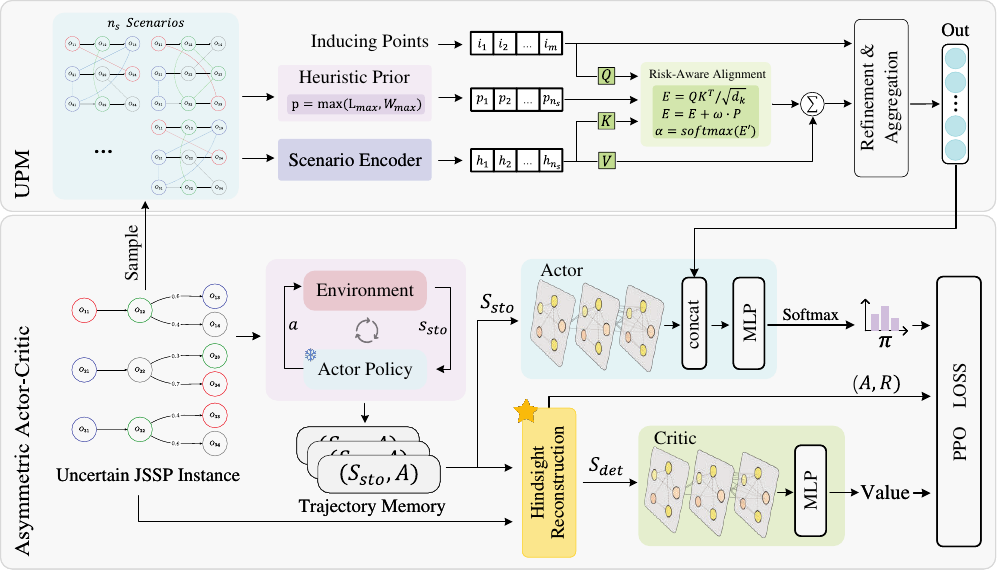}
    \caption{Overview of UP-AAC}
    \label{fig:overview}
\end{figure*}

\section{Preliminaries and Problem Formulation}

\subsubsection{Classical Job-Shop Scheduling Problem}
A classical JSSP instance is defined by a set of jobs $\mathcal{J} = \{J_1, \dots, J_n\}$ and a set of machines $\mathcal{M} = \{M_1, \dots, M_m\}$. Each job $J_i$ consists of a predetermined sequence of operations, where each operation requires a specific machine for a given processing time. The objective is to determine the start time for each operation to minimize the makespan, $C_{\max}$, subject to precedence and resource constraints.

\subsubsection{JSSP with Structural Uncertainty}
We extend the classical formulation to the JSSP with structural uncertainty. In this setting, an instance $\mathcal{I}$ still comprises a set of jobs $\mathcal{J}$ and machines $\mathcal{M}$, but the processing path for each job is not a fixed sequence. Instead,  it is defined as a probabilistic structure modeled by a Directed Acyclic Graph (DAG), $G_i=(O_i, E_i, P_i)$, where $O_i$ is the set of all potential operations for job $J_i$, $E_i$ represents feasible transitions between operations, and $P_i$ defines the conditional probability $P(o'|o)$ of transitioning from operation $o$ to operation $o'$. A scenario, denoted by $\omega$, is a single realization  in which a deterministic processing route is sampled for each job according to the corresponding transition probabilities. Each scenario corresponds to a standard JSSP instance, and the collection of all such scenarios defines the sample space.

\subsubsection{Standard Actor-Critic Framework}
Actor-Critic (AC) is a standard reinforcement learning paradigm \citep{Mnih2016AsynchronousMF}. It features two components: an Actor, which is a policy network $\pi_{\theta}(a|s)$ that maps a state $s$ to an action $a$, and a Critic, which is a value network $V_{\phi}(s)$ that estimates the expected return from that state. The Critic's role is to reduce the variance of the policy gradient by providing a value baseline. It does so by
computing the Advantage Function:
\begin{equation}
    A(s_t, a_t) = r_t + \gamma V_\phi(s_{t+1}) - V_\phi(s_t)
\end{equation}

The Actor updates using the advantage signal, while the Critic learns state values. However, high environmental stochasticity undermines the Critic’s ability to provide low-variance estimates.

\section{Methodology}

In this section, we present our end-to-end deep reinforcement learning framework, UP-AAC, designed to solve the JSSP under structural uncertainty. An overview of the framework is illustrated in Figure~\ref{fig:overview}. We begin by formulating the scheduling process as a Markov Decision Process (MDP). Subsequently, we introduce the two core components of our framework: first, the Asymmetric Actor-Critic (AAC) architecture, which is fundamental for ensuring stable learning in the stochastic environment; and second, the Uncertainty Perception Model (UPM), which enhances policy robustness by providing the agent with global risk insights. Finally, we outline the overall training procedure.

\subsection{Markov Decision Process Formulation}

We formulate the sequential decision-making process as an MDP. This allows us to use reinforcement learning to learn a dispatching policy $\pi$ that aims to minimize the makespan. The MDP is defined as follows.

\subsubsection{State ($\mathcal{S}$)}
The state $s_t \in \mathcal{S}$ is represented by a disjunctive graph, following the formulation by \citep{Baewicz2000TheDG}. Importantly, the representation is dynamic: it only includes information relevant to future decisions by excluding completed operations. Consequently, the state space shrinks as the scheduling process unfolds, improving computational efficiency. 

The state at step \(t\) is represented by a graph \(G_t\). To emphasize its correspondence to the stochastic and uncertain nature of the problem, we refer to it as the \textbf{stochastic state} (\(s_{sto}\)). The nodes \(\mathcal{O}_t\) in this graph correspond to all operations not yet completed, with each node featuring attributes that describe its current status (e.g., ongoing, ready), processing time, and relevant job-level metrics. The graph's structure is defined by two types of edges: directed conjunctive edges \(C_t\), which enforce precedence constraints within each job, and disjunctive edges \(D_t\), which represent the resource constraints among operations competing for the same machine.

\subsubsection{Action ($\mathcal{A}$)}
The agent makes decisions at discrete event points, typically when machines become idle and operations are ready. The action space $\mathcal{A}(s_t)$ consists of all eligible operations, defined as those whose predecessors have been completed. An action $a_t \in \mathcal{A}(s_t)$ corresponds to selecting one such operation for dispatch.

\subsubsection{State Transition ($P$)}
Upon taking an action $a_t$ in state $s_t$, the environment transitions to a new state $s_{t+1}$. This transition is inherently stochastic, driven by the structural uncertainty of the problem. For instance, if $a_t$ is a branching operation with multiple potential outcomes, the environment's dynamics will realize a single feasible path. This process renders the unchosen paths infeasible, fundamentally altering the graph topology for the subsequent state $s_{t+1}$. Once the dispatched operation $a_t$ is finished, it is removed from the active set.

\subsubsection{Reward ($R$)}
The objective is to minimize the final makespan $C_{\max}$. We employ a dense reward shaping mechanism, where the immediate reward $r_t$ is the negative growth of the estimated makespan lower bound, a technique proven effective in prior work \citep{zhang2020learning}.

\begin{equation}
    r(s_t, a_t) = C_{LB}(s_t) - C_{LB}(s_{t+1})
\end{equation}

Here, $C_{LB}(s)$ represents the makespan lower bound at state $s$, which is calculated as the length of the critical path in the disjunctive graph, considering only the precedence constraints and the processing times of the remaining operations. Since the initial lower bound $C_{LB}(s_0)$ is constant and the final bound $C_{LB}(s_T)$ equals the actual makespan, maximizing the cumulative reward is equivalent to minimizing the makespan.

\subsection{Asymmetric Actor-Critic}

While the standard AC framework is powerful, it encounters a fundamental challenge when applied to JSSP with structural uncertainty. The core issue lies in the high variance of the learning signal. In a standard AC setting, the Critic evaluates the stochastic state $s_{sto}$ to provide a baseline for the Actor's policy update. However, in our problem, the final outcome (and thus the cumulative reward) is not only determined by the quality of the dispatching actions, but also by the stochastic path realizations that follow each decision. A well-chosen action might still yield a poor makespan simply because the subsequent sequence of path choices results in a time-consuming path, and vice versa. This disconnect between action quality and eventual outcome introduces noise into the reward signal, making it unreliable for training the Critic. As a result, the value function becomes unstable, leading to high-variance advantage estimates and misleading gradients that hinder the Actor’s learning.

\begin{figure}[t]
    \centering
    \includegraphics[trim=0pt 0pt 0pt 10pt, clip, width=\columnwidth]{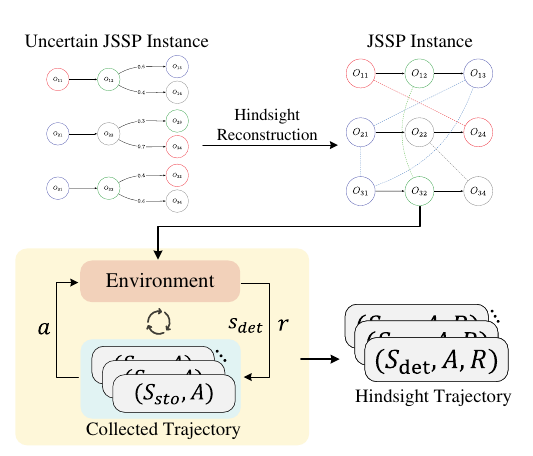}
    \caption{Hindsight Reconstruction}
    \label{fig:hindsight}
\end{figure}

To overcome this critical challenge of variance reduction, we propose an AAC architecture. The central idea of AAC is to decouple the Actor's exploration in a stochastic environment from the Critic's evaluation in a deterministic one. This is achieved through a mechanism we term \textbf{Hindsight Reconstruction}, as illustrated in Figure~\ref{fig:hindsight}.

The learning process under AAC involves two distinct phases. Initially, the Actor interacts with the stochastic environment, generating a trajectory of stochastic states and actions, $\tau_{sto} = (s_{sto,0}, a_0, s_{sto,1}, a_1, \dots)$. Once an episode terminates, the complete realized path of each job becomes known. We leverage this outcome to reconstruct a corresponding standard JSSP instance where all job routes are fixed and deterministic. By replaying the collected actions within this hindsight-reconstructed environment, we obtain a new trajectory composed of \textbf{deterministic states ($s_{det}$)} and rewards, $\tau_{det} = (s_{det,0}, a_0, r_0, s_{det,1}, a_1, r_1, \dots)$.

This asymmetric design assigns distinct roles and levels of informational access to the Actor and the Critic:
\begin{itemize}
    \item The \textbf{Actor} operates exclusively on the stochastic state $s_{sto}$. Its role is to learn a policy $\pi_{\theta}(a_t | s_{sto,t})$ that can navigate the uncertain environment, making robust decisions in the face of unknown future path realizations.
    \item The \textbf{Critic}, in contrast, is trained exclusively on the deterministic state $s_{det}$. Its purpose is to learn a state-value function $V_{\phi}(s_{det,t})$ that accurately estimates the expected return from a given state after the uncertainty has been resolved. Since the trajectory from any $s_{det,t}$ to the final outcome is fixed, the value estimation is conditioned on a deterministic future, thereby producing a stable and low-variance value baseline.
\end{itemize}

The Critic is updated by minimizing the mean squared temporal-difference error using the deterministic trajectory data. The Actor's policy gradient is then calculated using a low-variance advantage estimate, where the value baseline is provided by the Critic based on the corresponding deterministic state:

\begin{equation}
    A(s_{sto,t}, a_t) = r_t + \gamma V_{\phi}(s_{det, t+1}) - V_{\phi}(s_{det,t})
\end{equation}

By providing the Actor with a stable learning signal insulated from the stochasticity of the environment, our AAC framework effectively addresses the credit assignment problem and enables robust convergence towards a high-performance scheduling policy.

\subsection{Uncertainty Perception Model}
While the AAC framework provides a stable learning signal, the Actor's policy still mainly reacts to the current state $s_{sto}$. To equip the agent with a proactive, forward-looking perspective on a problem's intrinsic uncertainty, we introduce the Uncertainty Perception Model (UPM), which performs a one-time, offline analysis to generate a global risk feature vector, $z_{upm}$, to guide the Actor.

The UPM works based on a knowledge-guided learning principle. It starts by sampling to create a set of $n_s$ deterministic JSSP scenarios. Critically, instead of relying on a pure end-to-end approach, we inject explicit domain knowledge by first calculating a \textbf{heuristic risk prior} for each scenario $k$. This prior is based on two well-known scheduling metrics: maximum job length ($L_{\max}$) and maximum machine workload ($W_{\max}$) \citep{Pinedo1994SchedulingTA}, and serves as an efficient, interpretable measure of a scenario’s difficulty:

\begin{equation}
    p_k = \max(L_{\max}(\mathcal{SC}_k), W_{\max}(\mathcal{SC}_k))
\end{equation}

These priors then guide a risk-aware attention mechanism, which combines this domain knowledge with learned deep features \citep{Vaswani2017AttentionIA}. Each scenario $\mathcal{SC}_k$ is first encoded into a feature vector $h_k$ by a shared GNN encoder. To efficiently summarize information from the large set of scenarios, we use a small, fixed set of learnable inducing points $\{i_1, \dots, i_M\}$ as queries \citep{Lee2018SetT}. The attention score between an inducing point $i_j$ and a scenario feature $h_k$ is then modulated by the corresponding risk prior:

\begin{equation}
    e_{jk} = \frac{(W_q i_j)^T (W_k h_k)}{\sqrt{d_k}} + \omega \cdot p_k
\end{equation}

where $W_q, W_k$ are learnable projection matrices, and $\omega$ is a learnable scalar controlling the prior's influence. By incorporating the risk prior as a bias term, we explicitly guide the model to focus on scenarios identified as structurally challenging. This attention mechanism then aggregates information from all scenarios into the final risk vector $z_{upm}$. During online scheduling, this static $z_{upm}$ is subsequently concatenated with features from the dynamic state $s_{sto,t}$, thus effectively conditioning the Actor's policy, $\pi_{\theta}(a_t | s_{sto,t}, z_{upm})$, on both the immediate state representation and the  overall risks of the problem.

\begin{algorithm}[tb]
\caption{UP-AAC Training Procedure}
\label{alg:training_procedure}
\begin{algorithmic}[1]
\STATE \textbf{Initialize} Actor network $\pi_{\theta}$, Critic network $V_{\phi}$.
\FOR{each training iteration}
    \STATE Sample an uncertain JSSP instance $\mathcal{I}$
    \STATE \textit{// Phase 1: Uncertainty Perception}
    \STATE $z_{upm} \gets \text{UPM}(\mathcal{I})$
    
    \STATE \textit{// Phase 2:  Stochastic Trajectory Collection}
    \STATE Initialize a temporary buffer $\mathcal{B}_{temp} \leftarrow \emptyset$
    \FOR{$k=1$ \textbf{to} $K$}
        \STATE Collect trajectory $\tau_k \sim \pi_{\theta}(\cdot | s_{sto}, z_{upm})$
        \STATE Store $\tau_k$ in $\mathcal{B}_{temp}$
    \ENDFOR

    \STATE \textit{// Phase 3: Hindsight Reconstruction}
    \STATE Initialize an update buffer $\mathcal{B}_{update} \leftarrow \emptyset$
    \FOR{each trajectory $\tau_k$ in $\mathcal{B}_{temp}$}
        \STATE $(\{s_{det,t}\}, \{r_t\}) \gets \text{HindsightReconstruction}(\tau_k)$
        \STATE Store full transitions in $\mathcal{B}_{update}$
    \ENDFOR
    
    \STATE \textit{// Phase 4: Asymmetric Network Update}
    \STATE Advantages $A_t \gets r_t + \gamma V_{\phi}(s_{det,t+1}) - V_{\phi}(s_{det,t})$
    \STATE Update critic parameters $\phi$ using $\{s_{det,t}, r_t\}$
    \STATE Update actor parameters $\theta$ using $\{(s_{sto,t}, a_t, A_t)\}$
\ENDFOR
\end{algorithmic}
\end{algorithm}

\subsection{Training Procedure}
The overall training procedure of our UP-AAC framework integrates the offline uncertainty analysis from the UPM with the online policy learning of the AAC architecture. The process alternates between collecting experience in the stochastic environment and performing asymmetric updates using hindsight-reconstructed data. This workflow is detailed in Algorithm~\ref{alg:training_procedure}.

\section{Experiments}

\begin{table*}[!t]
\centering
\setlength{\tabcolsep}{1mm}
\begin{tabular}{cc|ccccccc|cc|c}
\toprule
\multicolumn{2}{c|}{\textbf{Instance}} & \textbf{FIFO} & \textbf{LOR} & \textbf{LWKR} & \textbf{LPT} & \textbf{MOR} & \textbf{MWKR} & \textbf{SPT} & \textbf{AC} & \textbf{UP-AAC} & \textbf{OR-Tools} \\
\midrule
\multirow{3}{*}{(5$\times$10, 1)} & Avg & 753.87 & 777.75 & 778.66 & 760.49 & 742.40 & 736.00 & 751.73 & 701.90 & \textbf{689.26} & 676.66 \\
& CVaR & 800.31 & 828.98 & 827.83 & 806.16 & 779.04 & 773.26 & 791.51 & 741.66 & \textbf{721.99} & 698.34 \\
& Gap  & 11.41\% & 14.94\% & 15.08\% & 12.39\% & 9.71\% & 8.77\% & 11.09\% & 3.73\% & \textbf{1.86\%} & 0.00\% \\
\midrule
\multirow{3}{*}{(5$\times$15, 2)} & Avg  & 1010.24 & 1031.48 & 1040.11 & 1022.65 & 998.56 & 992.53 & 1005.54 & 974.55 & \textbf{932.77} & 921.69 \\
& CVaR & 1084.38 & 1122.54 & 1133.86 & 1106.02 & 1069.75 & 1060.15 & 1080.05 & 1033.49 & \textbf{996.32} & 966.77 \\
& Gap & 9.61\% & 11.91\% & 12.85\% & 10.95\% & 8.34\% & 7.69\% & 9.10\% & 5.74\% & \textbf{1.20\%} & 0.00\% \\
\midrule
\multirow{3}{*}{(5$\times$20, 3)} & Avg & 1280.99 & 1296.44 & 1299.51 & 1295.80 & 1275.38 & 1274.97 & 1280.44 & 1256.09 & \textbf{1204.49} & 1191.56 \\
& CVaR & 1376.92 & 1403.05 & 1408.94 & 1395.66 & 1370.10 & 1367.95 & 1380.18 & 1341.90 & \textbf{1282.87} & 1252.64 \\
& Gap & 7.51\% & 8.80\% & 9.06\% & 8.75\% & 7.04\% & 7.00\% & 7.46\% & 5.42\% & \textbf{1.08\%} & 0.00\% \\
\midrule
\multirow{3}{*}{(10$\times$10, 1)} & Avg  & 1007.83 & 1102.92 & 1099.43 & 1091.29 & 964.60 & 955.89 & 1010.41 & 930.98 & \textbf{846.26} & 819.13 \\
& CVaR & 1065.04 & 1185.32 & 1180.06 & 1168.01 & 1016.19 & 1006.59 & 1078.95 & 975.54 & \textbf{943.73} & 845.72 \\
& Gap  & 23.04\% & 34.64\% & 34.22\% & 33.22\% & 17.76\% & 16.69\% & 23.35\% & 13.65\% & \textbf{3.31\%} & 0.00\% \\
\midrule
\multirow{3}{*}{(10$\times$15, 2)} & Avg  & 1265.94 & 1361.72 & 1365.95 & 1345.02 & 1220.62 & 1217.51 & 1257.94 & 1181.51 & \textbf{1079.24} & 1050.19 \\
& CVaR & 1359.53 & 1486.84 & 1495.74 & 1463.71 & 1293.85 & 1291.22 & 1357.74 & 1244.51 & \textbf{1191.87} & 1089.42 \\
& Gap & 20.54\% & 29.66\% & 30.07\% & 28.07\% & 16.23\% & 15.93\% & 19.78\% & 12.50\% & \textbf{2.77\%} & 0.00\% \\
\midrule
\multirow{3}{*}{(10$\times$20, 3)} & Avg  & 1529.60 & 1605.62 & 1614.16 & 1584.19 & 1489.93 & 1483.29 & 1516.66 & 1445.08 & \textbf{1334.83} & 1303.98 \\
& CVaR & 1635.01 & 1754.25 & 1765.32 & 1718.96 & 1580.98 & 1575.13 & 1632.05 & 1516.43 & \textbf{1457.83} & 1355.03 \\
& Gap  & 17.30\% & 23.13\% & 23.79\% & 21.49\% & 14.26\% & 13.75\% & 16.31\% & 10.82\% & \textbf{2.37\%} & 0.00\% \\
\bottomrule
\end{tabular}
\caption{Results on small and medium instances. The best performance among the competing methods is shown in \textbf{bold}.}
\label{tab:small_medium_results}
\end{table*}

In this section, we present a series of experiments to evaluate the performance of the proposed UP-AAC framework. Our evaluation is guided by two key questions: (1) How does UP-AAC compare to traditional heuristics and a standard reinforcement learning baseline across different problem scales? (2) What are the individual contributions of the framework’s core components, the AAC and the UPM?

\subsection{Experimental Setup}

\subsubsection{Datasets}
We evaluate our method on 12 sets of benchmark instances, which are procedurally generated by adapting the rules from the well-known Taillard benchmark suite \citep{Taillard1993BenchmarksFB} to cover a wide range of scales and uncertainty levels. The problem size is determined by the number of jobs ($n_j$) and machines ($n_m$), with $n_j \in \{5, 10, 15, 20\}$ and $n_m \in \{10, 15, 20\}$. The degree of structural uncertainty is directly coupled with the number of machines: instances with 10, 15, and 20 machines have 1, 2, and 3 branching operations per job ($n_b$), respectively. For each of the configurations, we generate 50 unique test instances.

\subsubsection{Baselines}
To demonstrate the effectiveness of our approach, we compare UP-AAC against a variety of baseline methods. The first category consists of seven widely-used priority dispatching rules (PDRs): first-in-first-out (FIFO), shortest-processing-time (SPT), longest-processing-time (LPT), most-operations-remaining (MOR), least-operations-remaining (LOR), least-work-remaining (LWKR), and most-work-remaining (MWKR). To provide a fair deep reinforcement learning comparison, we also implement a standard AC agent, which uses the same GNN architecture as our approach but is trained with a conventional setup where the Critic receives the same stochastic state $s_{sto}$ as the Actor \citep{zhang2020learning}. Finally, to provide a high-quality benchmark for calculating the performance gap, we define two reference solutions. For Small and Medium instances (where $n_j \le 10$), this benchmark is the optimal makespan ($C_{opt}$) obtained from a Constraint Programming (CP) solver. For Large instances (where $n_j \in \{15, 20\}$), for which the CP solver cannot find an optimal solution within the time limit, the benchmark is the best makespan found among all tested methods.

\subsubsection{Evaluation Metrics}
To thoroughly assess the overall performance under uncertainty, we conduct 50 independent evaluation runs for each test instance, each with a different random seed. For a comprehensive evaluation, we report the following three key metrics. The average makespan (Avg) over all runs measures the general performance of a policy. To evaluate a policy's robustness, we use the Conditional Value-at-Risk (CVaR) of the makespan, calculated as the average of the worst 20\% of outcomes. Finally, we compute the percentage gap (Gap (\%)) relative to the high-quality baselines established in the previous section.

\subsubsection{Implementation Details}
We train a separate, specialized UP-AAC model for each of the 12 instance configurations. All models are implemented in PyTorch and trained on a single NVIDIA V100 GPU. The core GNN encoder consists of 3 graph convolution layers, with a hidden dimension of 72. The Actor and Critic networks are both 3-layer MLPs, which incorporate Layer Normalization \citep{Ba2016LayerN} and LeakyReLU activation functions \citep{maas2013rectifier} to enhance training stability. For the UPM module, we sample $n_s=100$ deterministic scenarios for each instance during training to construct the stochastic path model.

\begin{table*}[!t]
\centering
\setlength{\tabcolsep}{1mm} 
\begin{tabular}{cc|ccccccc|cc|c}
\toprule
\multicolumn{2}{c|}{\textbf{Instance}} & \textbf{FIFO} & \textbf{LOR} & \textbf{LWKR} & \textbf{LPT} & \textbf{MOR} & \textbf{MWKR} & \textbf{SPT} & \textbf{AC} & \textbf{UP-AAC} & \textbf{Best Found} \\
\midrule
\multirow{2}{*}{(15$\times$10, 1)} & Avg & 1241.13 & 1405.14 & 1408.37 & 1402.99 & 1190.53 & 1191.56 & 1254.51 & 1178.35 & \textbf{1161.13} & 1158.55 \\
& CVaR & 1314.84 & 1509.08 & 1507.69 & 1505.57 & 1248.51 & 1255.02 & 1341.12 & 1242.49 & \textbf{1215.98} & 1210.14 \\
\midrule
\multirow{2}{*}{(15$\times$15, 2)} & Avg & 1521.71 & 1700.57 & 1689.59 & 1673.76 & 1437.89 & 1444.96 & 1511.80 & 1429.96 & \textbf{1407.05} & 1404.42 \\
& CVaR & 1626.64 & 1854.18 & 1831.08 & 1812.22 & 1515.35 & 1524.47 & 1632.16 & 1507.37 & \textbf{1473.68} & 1468.70 \\
\midrule
\multirow{2}{*}{(15$\times$20, 3)} & Avg & 1791.33 & 1973.78 & 1986.03 & 1934.59 & 1708.08 & 1704.50 & 1784.34 & 1683.61 & \textbf{1670.13} & 1667.10 \\
& CVaR & 1913.04 & 2161.43 & 2165.06 & 2104.42 & 1799.46 & 1798.16 & 1928.85 & 1772.62 & \textbf{1748.38} & 1742.57 \\
\midrule
\multirow{2}{*}{(20$\times$10, 1)} & Avg & 1471.99 & 1704.39 & 1730.41 & 1682.47 & 1397.24 & 1424.95 & 1514.78 & 1415.82 & \textbf{1378.27} & 1376.08 \\
& CVaR & 1552.38 & 1822.66 & 1861.25 & 1802.64 & 1461.59 & 1492.85 & 1613.14 & 1506.21 & \textbf{1437.07} & 1431.56 \\
\midrule
\multirow{2}{*}{(20$\times$15, 2)} & Avg & 1766.18 & 2014.25 & 2007.34 & 1987.45 & 1666.12 & 1676.38 & 1773.33 & 1661.20 & \textbf{1639.89} & 1637.07 \\
& CVaR & 1877.36 & 2181.69 & 2176.71 & 2146.71 & 1750.42 & 1763.55 & 1904.00 & 1769.69 & \textbf{1716.68} & 1710.56 \\
\midrule
\multirow{2}{*}{(20$\times$20, 3)} & Avg & 2027.98 & 2275.24 & 2282.10 & 2236.04 & 1910.47 & 1925.53 & 2014.10 & 1922.53 & \textbf{1884.30} & 1881.06 \\
& CVaR & 2160.19 & 2463.33 & 2474.15 & 2414.41 & 2006.35 & 2020.64 & 2173.87 & 2060.13 & \textbf{1967.20} & 1960.58 \\
\bottomrule
\end{tabular}
\caption{Results on large instances. The best performance among the competing methods is shown in \textbf{bold}.}
\label{tab:large_results}
\end{table*}

\begin{table*}[!t]
\centering
\setlength{\tabcolsep}{1mm} 
\begin{tabular}{c|cc|cc|cc|cc|cc|cc}
\toprule
\multirow{2}{*}{\textbf{Method}} & \multicolumn{2}{c|}{\textbf{(5$\times$10, 1)}} & \multicolumn{2}{c|}{\textbf{(5$\times$15, 2)}} & \multicolumn{2}{c|}{\textbf{(5$\times$20, 3)}} & \multicolumn{2}{c|}{\textbf{(10$\times$10, 1)}} & \multicolumn{2}{c|}{\textbf{(10$\times$15, 2)}} & \multicolumn{2}{c}{\textbf{(10$\times$20, 3)}} \\
& Avg & CVaR & Avg & CVaR & Avg & CVaR & Avg & CVaR & Avg & CVaR & Avg & CVaR \\
\midrule
Standard AC & 701.90 & 741.66 & 974.55 & 1033.49 & 1256.09 & 1341.90 & 930.98 & 975.54 & 1181.51 & 1244.51 & 1445.08 & 1516.43 \\
w/o AAC & 709.41 & 745.60 & 970.57 & 1034.04 & 1252.70 & 1340.64 & 934.49 & 1000.31 & 1186.83 & 1268.58 & 1434.63 & 1515.75 \\
w/o UPM & 695.72 & 724.22 & 938.85 & 1008.61 & 1211.05 & 1296.13 & 857.50 & 963.27 & 1099.29 & 1226.72 & 1369.99 & 1508.02 \\
\textbf{UP-AAC} & \textbf{689.26} & \textbf{721.99} & \textbf{932.77} & \textbf{996.32} & \textbf{1204.49} & \textbf{1282.87} & \textbf{846.26} & \textbf{943.73} & \textbf{1079.24} & \textbf{1191.87} & \textbf{1334.83} & \textbf{1457.83} \\
\bottomrule
\end{tabular}
\caption{Ablation study results on Small and Medium instances.}
\label{tab:ablation_study}
\end{table*}

\subsection{Main Results Analysis}
The results of our experiments on all 12 instance sets are summarized in Table~\ref{tab:small_medium_results} (Small and Medium instances) and Table~\ref{tab:large_results} (Large instances). The data indicates a consistent performance advantage for the proposed UP-AAC framework across the tested scenarios.

Across all instance sizes and uncertainty levels, UP-AAC achieves the best performance among all competing methods. In terms of average makespan (Avg), it consistently yields the most efficient solutions. Notably, its advantage is even more pronounced in the Conditional Value-at-Risk (CVaR) metric. The lower CVaR values reflect the method's robustness, highlighting its ability to effectively mitigate risks and maintain stability in worst-case scenarios. This robustness is further evidenced by the small percentage gap to the optimal or best-known solutions—averaging 1.94\% on small/medium and 0.18\% on large instances—which underscores its capability to generate near-optimal schedules.

A comparison of the baseline methods provides further insights into the problem. The Priority Dispatching Rules (PDRs) exhibit a clear performance hierarchy, where heuristics prioritizing jobs with more remaining work (MOR and MWKR) consistently form the strongest baselines. This suggests that a forward-looking strategy is inherently more effective in environments with structural uncertainty. While the standard AC model outperforms all PDRs, confirming the potential of DRL, it still shows a significant performance gap compared to UP-AAC. This disparity highlights the limitations of conventional AC frameworks when faced with the high-variance learning signals inherent to our problem.

On large-scale instances (Table~\ref{tab:large_results}), while UP-AAC maintains its top-ranking performance, its relative improvement over the strongest baselines like MWKR appears to be smaller. This phenomenon can be attributed to two factors. First, as problem size and congestion increase, the scheduling environment becomes more constrained, making powerful heuristics that manage key bottlenecks more effective. UP-AAC’s ability to consistently find superior solutions in these environments demonstrates its strength in fine-grained optimization. Second, the Gap (\%) for large instances is calculated against the Best Found solution from the experiment itself, rather than a certified optimum. This internal baseline is inherently less stringent than the CP solver's optimal solution used for smaller instances, which can lead to a perceived reduction in the performance gap.

\subsection{Ablation Studies}

To validate the individual contributions of our two core components—the AAC architecture and the UPM—we conducted a series of ablation studies. We compare our full model (UP-AAC) against two variants: (1) \textbf{w/o UPM}, which removes the UPM module and relies solely on the AAC framework; and (2) \textbf{w/o AAC}, which replaces our asymmetric architecture with a standard AC setup while still utilizing the UPM. The results are summarized in Table~\ref{tab:ablation_study}.

The experimental results highlight the substantial contribution of the UPM. Removing the UPM leads to a consistent degradation in performance across all test instances. This decline is particularly evident in the CVaR metric compared to the average makespan. For instance, on the (10$\times$20, 3) instance, the average makespan increases by 2.6\%, while the CVaR increases by 3.4\%. These findings confirm that the global risk embedding is crucial for enhancing policy robustness and the ability to navigate worst-case scenarios.

The ablation study further reveals that the AAC architecture is a critical determinant of overall performance. Replacing the AAC module (w/o AAC) results in a pronounced reduction in performance; the w/o AAC model consistently underperforms the AAC-only variant and, in several instances, even fails to match the standard AC baseline. For example, on the (10$\times$10, 1) instance, the average makespan of the w/o AAC model is higher than that of the standard AC. These results validate our core hypothesis: in environments with high structural uncertainty, simply providing additional information (via UPM) to an unstable learning framework is insufficient and can even be detrimental. The stable, low-variance learning signal provided by AAC's hindsight reconstruction appears to be an essential prerequisite for advanced features to be effectively utilized by the agent.

\section{Conclusion}

In this paper, we tackled the Job-Shop Scheduling Problem under structural uncertainty, identifying the incorrect credit assignment issue as a core challenge for standard DRL methods. We introduced UP-AAC, a novel framework whose cornerstone is an Asymmetric Actor-Critic (AAC) architecture. By training the Critic on a deterministic state reconstructed in hindsight, AAC provides a stable learning signal that effectively resolves the credit assignment problem. This foundation is further enhanced by our Uncertainty Perception Model (UPM), which injects a knowledge-guided global risk assessment to improve policy robustness. Extensive experiments have validated that UP-AAC significantly outperforms strong baselines in both average performance and risk mitigation. We believe the principle of hindsight reconstruction within our asymmetric framework offers a promising methodology for other combinatorial optimization problems facing similar uncertainty. Furthermore, the UPM’s scenario-based design can be extended to leverage historical production data, paving the way for continuously learning industrial scheduling systems.


\section*{Acknowledgments}

This work was supported in part by the National Natural Science Foundation of China under Grant Nos. 62372028, 62372027, and U23B2025; and by a CAHSI–Google Institutional Research Program award.


\bibliography{aaai2026}

\end{document}